\documentclass[conference]{IEEEtran}
\IEEEoverridecommandlockouts
\usepackage{cite}
\usepackage{amsmath,amssymb,amsfonts}
\usepackage{algorithmic}
\usepackage{graphicx}
\usepackage{textcomp}
\usepackage{xcolor}
\def\BibTeX{{\rm B\kern-.05em{\sc i\kern-.025em b}\kern-.08em
    T\kern-.1667em\lower.7ex\hbox{E}\kern-.125emX}}
\begin{document}

\title{Words that Represent Peace\\
\thanks{Funding provided by the Climate School and Data Science Institute at Columbia University for student researchers and overall support provided by the Toyota Research Institute.}
}

\author{\IEEEauthorblockN{Tushar Prasad}
\IEEEauthorblockA{\textit{Data Science Institute} \\
\textit{Columbia University}\\
New York, NY, USA \\
tp2802@columbia.edu}

\and
\IEEEauthorblockN{ Larry S. Liebovitch}
\IEEEauthorblockA{\textit{AC4 in the Climate School} \\
\textit{Columbia University}\\
New York, NY, USA \\
lsl2140@columbia.edu}

\and
\IEEEauthorblockN{Melissa Wild}
\IEEEauthorblockA{\textit{AC4 in the Climate School} \\
\textit{Columbia University}\\
New York, NY, USA \\
mm3484@columbia.edu}

\and
\IEEEauthorblockN{Harry West}
\IEEEauthorblockA{\textit{Indust.l Engr. \& Ops. Research} \\
\textit{Columbia University}\\
New York, NY, USA \\
hw2599@columbia.edu}

\and
\IEEEauthorblockN{Peter T. Coleman}
\IEEEauthorblockA{\textit{Teachers College \& AC4 in the Climate School} \\
\textit{Columbia University}\\
New York, NY, USA \\
coleman@exchange.tc.columbia.edu}

}

\maketitle

\begin{abstract}
We used data from LexisNexis to determine the words in news media that best classifies countries as higher or lower peace.  We found that higher peace news is characterized by themes of finance, daily actitivties, and health and that lower peace news is characterized by themes of politics, government, and legal issues.  This work provides a starting point to measure levels of peace and identify the social processes that underly those words.
\end{abstract}


\section{Introduction}
Communication through language is not just a medium for exchanging information, but a powerful tool that shapes our perception of reality and plays a pivotal role in the dynamics of human interactions, including conflicts. The destructive potential of "hate speech" in inciting violence and deepening societal divisions is well-documented \cite{ezeibe}\cite{peacetech}. Simultaneously, peacekeepers and researchers are leveraging data science and natural language processing (NLP) to monitor and predict such conflicts, aiming to mitigate the escalation of hostilities through proactive measures \cite{diehl}.\\

However, while much of the research has been focused on the prevention of conflict, there is a growing interest in understanding the conditions that foster sustained peace \cite{fry}\cite{ampsych}. Highly peaceful societies are characterized by more than just the absence of violence; they exhibit specific norms, values, and stability that reduce the likelihood of conflict re-emergence. This has led to an increased focus on "positive peace" – a concept that seeks to identify and understand the active social forces that contribute to the harmonious coexistence within societies.\\

Against this backdrop, this paper explores the linguistic dimensions of peace and conflict across various levels of language structure, including phonology, grammar, semantics, pragmatics, and discourse. We specifically investigate "peace speech," a concept that could be considered the counterpart of "hate speech." Peace speech involves linguistic structures that promote and sustain peaceful interactions among individuals and groups. Despite its significance, empirical studies on the specific features and effects of peace speech are scarce. \\

This study employs machine learning techniques to identify key linguistic features of peace speech. By analyzing the most frequently used words in countries with varying levels of peace, we aim to uncover insights into how language can be harnessed to foster a more peaceful society. Through this research, we seek to move from theoretical possibilities to empirical evidence, providing a clearer understanding of the linguistic underpinnings that support peaceful environments.\\

In this complex landscape, journalists play a crucial role in either propagating paranoia or fostering peace through the themes and social processes they choose to highlight in their reports. The narratives constructed in newspapers and journals can significantly influence a society's perception of its peace or turmoil. Imagine a system that could inform journalists of the "peacefulness" of the society they are writing about and offer suggestions to adjust their articles to promote peace while delivering the same core message. Such a system would be invaluable in steering public discourse toward constructive ends. Our research aims to lay the groundwork for this by starting with an analysis of how words can define and influence the peacefulness of a society, seeking to develop a deeper understanding of the linguistic elements that make a difference in these contexts.

\section{Related Works}

According to recent research led by Fry et al. \cite{fry} and Coleman et al. \cite{ampsych}, the study of societal peacefulness reveals that many groups worldwide actively choose peace over war, representing a largely untapped resource for delineating pathways to peace. The findings suggest that although the conditions for sustainable peace are complex and vary greatly between societies, they can be broadly understood through the ratio of positive to negative intergroup reciprocity, which remains stable over time. This core dynamic has facilitated the connection of diverse studies across multiple disciplines, providing a more unified and comprehensive view of how peace is sustained. Moreover, the research emphasizes the importance of local contexts in interpreting key variables of peacefulness, as exemplified by the varied impacts of religiosity in different societies. This nuanced approach underscores the potential of using scientific methods to celebrate and learn from the multitude of peaceful communities around the globe.\\

Liebovitch et al. \cite{lsl} illustrate how peace extends beyond the absence of conflict, defining "positive peace" as the systems that foster and sustain harmonious societies. Their work utilizes complex systems analysis, inspired by physics, to explore the interplay of factors contributing to peace. By developing causal loop diagrams and differential equation models, they identify key attractors and dynamics within these systems. Their approach incorporates data science techniques to quantify peace factors from diverse data sources, including social media, and introduces a graphical interface allowing stakeholders to simulate changes in these systems. This comprehensive analysis demonstrates the effectiveness of methods traditionally used in physical and biological sciences for understanding the complex dynamics essential for sustainable peace. \\

Fry et al. \cite{fry} provide a comprehensive analysis of peace systems, demonstrating that some human societies not only avoid war but also actively foster positive intergroup relationships. Their research identifies peace systems as clusters of neighboring societies that maintain peace among each other, highlighting the feasibility of creating harmonious intergroup relations across different social units, from tribal societies to nations. This study significantly contributes to our understanding by showing that peace systems exhibit higher levels of common identity, interconnectedness, interdependence, and peace leadership compared to other social systems. Furthermore, through machine learning analysis, Fry et al. discovered that non-warring norms, rituals, and values are crucial for maintaining peace within these systems. These findings have profound implications for policy-making, suggesting that promoting characteristics of peace systems can enhance international cooperation to tackle global challenges such as pandemics, environmental issues, and nuclear proliferation.\\

Liebovitch et al. \cite{plosone} explored the linguistic differences in news media between countries with varying levels of peace using natural language processing and machine learning. Their study revealed that language plays a crucial role as both a cause and a consequence of social processes leading to peace or conflict. By analyzing online news sources and utilizing existing peace indices, they identified key words that distinguish lower-peace countries from higher-peace countries. Instead of starting with a theoretical framework to predict which words would be more prevalent in different peace contexts, they used machine learning to empirically identify words that most accurately classified a country's peace status. The study also developed a machine learning model trained on word frequencies from countries at the extremes of peace. This model was then applied to compute a quantitative peace index for intermediate-peace countries, effectively bridging the gap between lower-peace and higher-peace. This approach demonstrates the potential of natural language processing and machine learning to generate new quantitative measures of social systems, in this case, providing a nuanced linguistic perspective on peace across various countries.

\section{Methodology}

\subsection{Data}

For our analysis of peace speech versus hate speech in media, we utilized a new,  comprehensive dataset provided by LexisNexis through a partnership with Elsevier and Columbia University. This dataset encompasses approximately 2,000,000 media articles from 2010 to 2020, written in English, from 20 different countries. The variety of sources includes publications such as "24 Hours Toronto," "BBC Monitoring: International Reports," and "Marie Claire," among others.\\

\subsubsection{Measuring Peace}

We have used the peace studies listed below to train our model. We utilize an average of the values from these indices to determine which countries are classified as peaceful and which are non-peaceful.

\begin{itemize}
    \item Global Peace Index \cite{gpi}
    \item Positive Peace Index \cite{ppi}
    \item Human Development Index \cite{hdi}
    \item World Happiness Index \cite{whi}
    \item Fragile States Index \cite{fsi}
    \item Inclusiveness Index \cite{ii}
    \item Gini Coefficient \cite{gini}\\
\end{itemize}

From this extensive collection, we extracted and analyzed 10,000 words from media outlets in both peaceful and non-peaceful countries, as defined by the Indexes above. The peaceful countries in our study were Austria, Australia. Belgium, Czech Republic, Denmark, Finland, Netherlands, New Zealand, Norway, and Sweden; while the non-peaceful countries were Afghanistan, Congo, Guinea, India. Iran, Kenya, Nigeria, Sri Lanka, Uganda and Zimbabwe. This distinction allowed us to compare the linguistic features of peace speech and hate speech across different societal contexts and development levels, focusing on articles written in English rather than in the local languages of the countries involved.

\subsection{Preprocessing}

To ensure the robustness and relevance of our analysis of peace speech versus hate speech in media articles, we followed a series of preprocessing steps to refine our dataset and prepare it for machine learning modeling:\\

\subsubsection{Extraction of Word Occurrences}

We started by extracting words and their number of occurrences from the media articles for each country. This initial extraction focused on words that were relevant according to the Human Development Indexes, identifying the frequency of each word within the articles from both peaceful and non-peaceful countries.\\

\subsubsection{Removal of Proper Nouns}

To achieve unbiased results and prevent specific entities or names from skewing the analysis, we removed proper nouns from our dataset. This step ensures that common nouns, adjectives, and verbs primarily drive the patterns identified rather than specific names that could disproportionately influence the results.\\

\subsubsection{Selective Stop Word Removal}

After consulting with linguistic and subject-matter experts, we removed a significant number of stop words. However, this was done selectively; articles, auxiliary verbs, conjunctions, and prepositions were primarily targeted for removal. We retained some stop words we believed could reveal meaningful patterns in the linguistic features of peaceful versus non-peaceful societies. This nuanced approach allowed us to maintain potentially insightful linguistic markers while eliminating noise.\\

\subsubsection{Reduction to Top Words}

To further refine our dataset, we reduced the number of words per country to 1,000. This reduction was based on the rationale that a larger corpus might include rare words, leading to model overfitting, while a too-small corpus could result in significant information loss. By choosing 1,000 words, we struck a balance, keeping the most relevant words that are reasonably frequent across the corpus.\\

\subsubsection{Normalization of Word Occurrences}

The next step was to normalize the word frequencies to account for variations in total word counts across different countries. This normalization was crucial because some countries had disproportionately high numbers of words, which could bias the analysis if not adjusted. For each country, we computed the total number of occurrences of all words, $N(country)$. The normalized word count for each word in that country was then calculated as

\begin{equation}
    \text{Normalized Word Count} = \frac{N(\text{word, country})}{N(\text{country})}
\end{equation}

After normalization, for each word, we calculated the average of these normalized word frequencies across all countries within the peaceful and non-peaceful datasets separately. This average:

\begin{equation}
    \text{Avg Normalized Frequency} = \frac{\sum (\text{Normalized Word Count})}{\text{Number of Countries}}
\end{equation}

By following these preprocessing steps, including crucial normalization to handle disproportionate word counts, we prepared the dataset to effectively highlight the linguistic differences between peaceful and non-peaceful societies. This sets the stage for a deeper machine-learning analysis to identify and understand the key features of peace speech.

\subsection{Modeling}

After preprocessing, we concatenated the normalized data for peaceful and non-peaceful countries to prepare for the machine-learning phase. To explore the linguistic differences that might predict a country's peace status, we employed four different supervised learning techniques: Logistic Regression, Support Vector Machines (SVMs), Decision Trees, and Random Forests. Each of these models was selected for its unique strengths in classification tasks, allowing us to cross-verify the insights gained and ensure robustness in our findings.\\

Given that our dataset comprised data from only twenty countries, traditional methods of cross-validation were not feasible without risking overfitting. Instead, we adopted a novel approach inspired by the leave-one-out cross-validation technique. In this method, we used one country at a time as the holdout set and trained our models on the remaining nineteen countries. This process was repeated for each country, thus each time predicting whether the held-out country was peaceful or non-peaceful based on its linguistic profile.\\

The primary metric used to evaluate the performance of our models was accuracy, which was chosen due to its intuitive interpretability in binary classification tasks. To calculate this, we averaged the accuracies obtained from each iteration of our leave-one-out approach, providing a comprehensive view of how well our models could generalize across diverse linguistic contexts. Additionally, we compiled the results into a confusion matrix for each model, allowing us to see not just the overall accuracy but also the specific instances of true positives, true negatives, false positives, and false negatives. This detailed breakdown helped in understanding the strengths and limitations of each modeling technique in the context of predicting peace status from linguistic features.

\subsection{Qualitative Inference}

\subsubsection{Word Clouds}

After obtaining the optimal models, we created two distinct word clouds to visually represent the linguistic characteristics of peaceful and non-peaceful countries. The size of each word within these clouds corresponds to its frequency of occurrence in the respective group. This visualization technique enabled us to identify and compare the prominent themes and terms that distinguish peaceful countries from their non-peaceful counterparts. The word clouds facilitated an intuitive exploration of the data, helping us to recognize the specific linguistic elements that are prevalent in different types of societies.\\

\subsubsection{Dimensionality Reduction}

After analyzing the word clouds and gaining a qualitative understanding of the distinct linguistic features associated with peaceful and non-peaceful countries, we sought to further investigate the relationships and similarities among the words. To do this, we first embedded the words using a large language model, which provided us with high-dimensional representations of each word's semantic meaning. Given the complexity and interpretability challenges associated with high-dimensional data, we employed Principal Component Analysis (PCA) as a method for dimensionality reduction. PCA enabled us to reduce the dimensionality of our word embeddings to a two-dimensional space, making it possible to visually interpret the relationships and clusters among the words.\\

\subsubsection{Clustering}

With the data transformed into a 2D space, we applied K-means clustering to identify distinct groups or clusters of words that shared similar characteristics. This step was crucial for manually examining how words from the word clouds clustered together, potentially revealing underlying patterns that were not immediately apparent from the word clouds alone. We plotted the results of the PCA and K-means clustering to visually inspect and analyze the clusters, providing a deeper insight into the semantic connections between the words used in the media across different countries.\\

\subsubsection{Semantic Segmentation using a Large Language Model}

In our study, we extended our analysis by utilizing a large language model (LLM) for semantic segmentation of the words collected from the word clouds. By feeding these words into the LLM, we asked it to independently segment them based on semantic similarities, aiming to validate our manual clustering results. This approach provided a robust comparison, testing whether human intuition aligned with an AI system recognized for its high accuracy in human-like text understanding. This dual-method analysis not only reinforced our findings but also explored the potential of AI in capturing nuanced linguistic relationships in the context of peace and conflict studies.

\section{Experiments and Results}

\subsection{Data and Preprocessing}

We selected the top 1,000 words across peaceful and non-peaceful countries separately, as outlined in the methodology, and subjected them to all the preprocessing steps detailed in the methodology. After preprocessing, we were left with 1,270 unique words aggregated across peaceful and non-peaceful sections. Since the aggregate occurrences between certain words after normalization were very far apart, we attempted to dampen the differences by log-transforming all the values.

\subsection{Modeling Experiments}

\begin{table}[htbp]
\caption{Model Evaluation}
\begin{center}
\begin{tabular}{|c|c|c|c|}
\hline 
\textbf{Model} & \textbf{\textit{Precision}}& \textbf{\textit{Recall}}& \textbf{\textit{Accuracy}} \\
\hline
Logistic Regression& 100\% &100\% & 100\% \\
\hline
SVM& 100\% & 100\%& 100\% \\
\hline
Decision Tree& 100\% &95\% & 95\% \\
\hline
Random Forest& 100\% & 95\%& 95\% \\
\hline
\end{tabular}
\label{tab1}
\end{center}
\end{table}

We used Logistic Regression, Support Vector Machines (SVMs), Decision Trees, and Random Forests models for the classification tasks. Following the methodology, we adopted a leave-one-out cross-validation approach, using one country at a time as the holdout set and training our models on the remaining nineteen countries. This method ensured that each country was tested for its peace status prediction based on its linguistic profile.\\

We performed hyperparameter tuning using the Optuna package to optimize the performance of each model. The primary metric used to evaluate the models was accuracy. The metric results are presented in the table I.\\

We observed very good results from all the models after hyperparameter tuning. Although Logistic Regression and SVMs showed higher accuracy, we decided to proceed with the SVM and Random Forest models for qualitative analysis. Since the hyperparameter-tuned SVM we obtained uses a linear kernel, we wanted to include variety in the qualitative analysis by also using the Random Forest model. This approach provides a broader perspective on the linguistic features of peaceful and non-peaceful countries. The detailed breakdown also informed this decision of true positives, true negatives, false positives, and false negatives provided by the confusion matrices for each model, which helped us understand the strengths and limitations of each technique in predicting peace status from linguistic features.

\subsection{Qualitative Results}

\subsubsection{Word Clouds}

\begin{figure}[htbp]
\centerline{\includegraphics[width=0.5\textwidth]{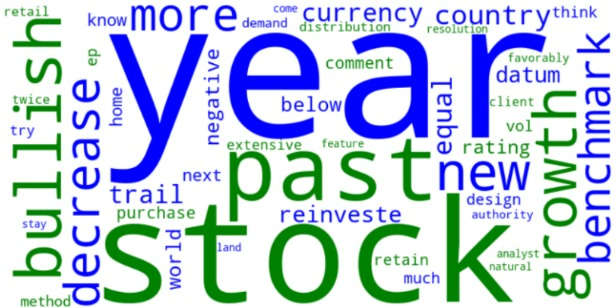}}
\caption{Random Forest Peaceful Word Cloud}
\label{fig1}
\end{figure}

\begin{figure}[htbp]
\centerline{\includegraphics[width=0.5\textwidth]{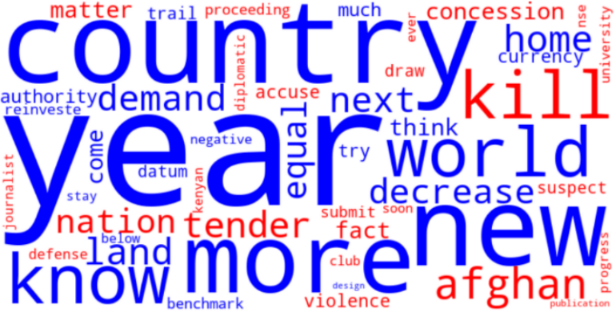}}
\caption{Random Forest Nonpeaceful Word Cloud}
\label{fig2}
\end{figure}

After obtaining the optimal models, we created word clouds to visually represent the linguistic characteristics of peaceful and non-peaceful countries shown in Figs. \ref{fig1} and \ref{fig2}. For the Random Forest model, we selected the top 50 words based on their feature importance. The size of each word in the word clouds corresponds to its frequency of occurrence in the respective group. Words appearing in both peaceful and non-peaceful countries are highlighted in blue, allowing for a direct comparison of shared linguistic elements.\\

\begin{figure}[htbp]
\centerline{\includegraphics[width=0.5\textwidth]{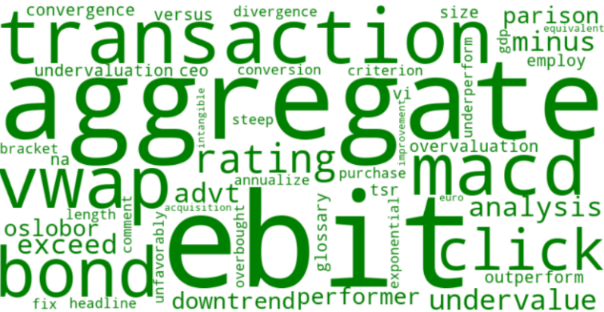}}
\caption{SVM Peaceful Word Cloud}
\label{fig3}
\end{figure}

\begin{figure}[htbp]
\centerline{\includegraphics[width=0.5\textwidth]{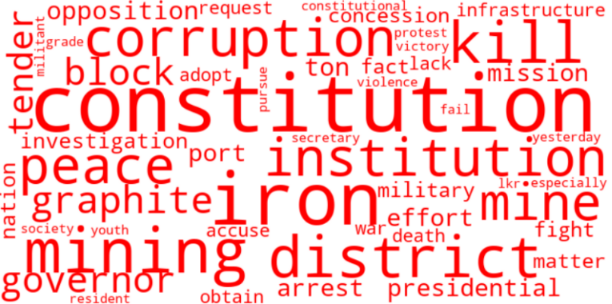}}
\caption{SVM Nonpeaceful Word Cloud}
\label{fig4}
\end{figure}

For the SVM model, we generated word clouds from the top 75 words contributing to the classification of peaceful and non-peaceful countries based on their coefficients shown in Figs. \ref{fig3} and \ref{fig4}. The size of each word reflects its frequency of occurrence in the respective group. This approach provided insights into the specific linguistic elements that are most influential in distinguishing peaceful and non-peaceful societies.\\

Both word clouds effectively highlighted the distinct linguistic landscapes between peaceful and non-peaceful societies. Peaceful countries are characterized by a discourse focused on economic stability, financial markets, and growth, as reflected in terms like "transaction," "aggregate," "year," and "stock." In contrast, non-peaceful countries are marked by discussions on violence, conflict, and political instability, with prominent terms such as "corruption," "kill," "constitution," and "afghan." These word clouds provide a clear and intuitive understanding of the specific linguistic elements that differentiate peace speech from hate speech, offering valuable insights into the underlying themes prevalent in different types of societies.\\

\subsubsection{Dimensionality Reduction and Clustering}

\begin{figure}[htbp]
\centerline{\includegraphics[width=0.5\textwidth]{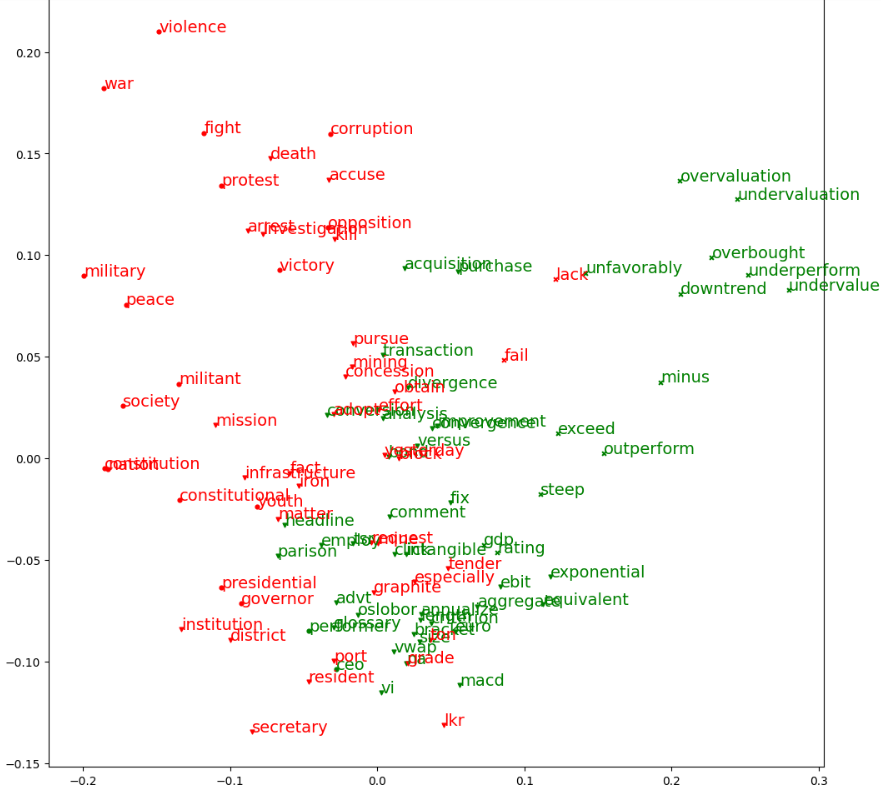}}
\caption{SVM K-means Clustering}
\label{fig5}
\end{figure}

\begin{figure}[htbp]
\centerline{\includegraphics[width=0.5\textwidth]{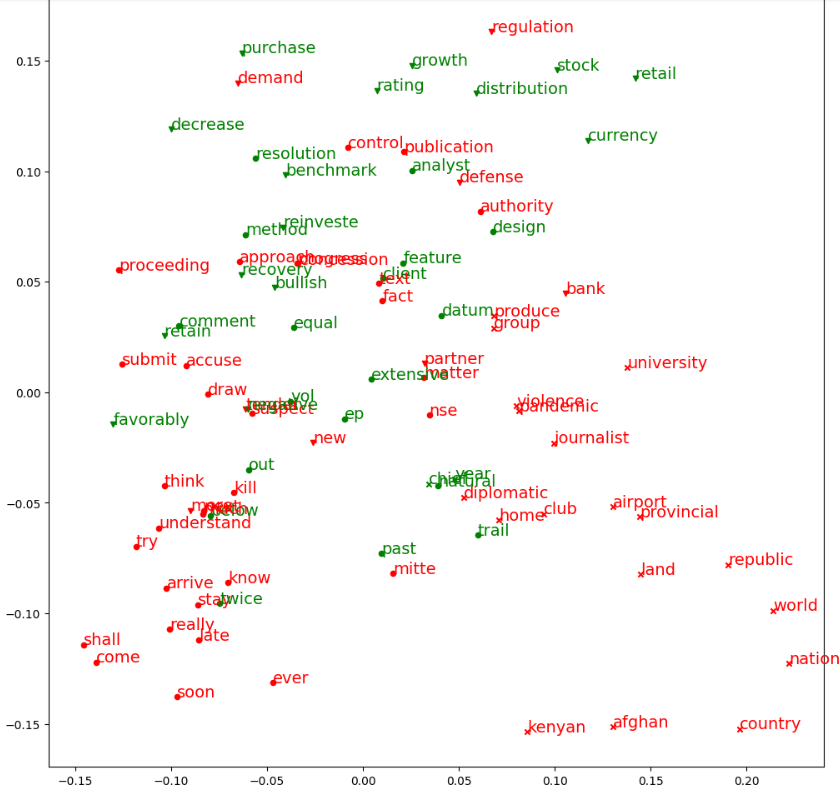}}
\caption{Random Forest K-means Clustering}
\label{fig6}
\end{figure}

We embedded the words using the multilingual-e5-large-instruct model, obtaining high-dimensional representations of each word’s semantic meaning. PCA allowed us to transform the word embeddings into a two-dimensional space, facilitating visual interpretation of the relationships and clusters among the words. With the data transformed into a 2D space, we applied K-means clustering to identify distinct groups or clusters of words that shared similar characteristics, as shown in Figs. \ref{fig5} and \ref{fig6} In our analysis, we performed K-means clustering with three clusters for both the SVM and Random Forest word clouds. For the SVM model, the identified themes included: Warfare, political strife, or news reporting on conflicts; Stock Market Performance and Company Performance Reviews; and Business, Economics, Governance, and Policy-making. For the Random Forest model, the clusters reflected themes of: Processes, Analytical methods, Legal and Investigative actions; Cultural, Political, and Global Dynamics; and Economic and Financial Activities.\\

\subsubsection{AI Semantic Segmentation}

For the peaceful word clouds, the LLM segmented the words into several distinct themes. The Random Forest model's themes included Finance, Personal Social, Health Wellbeing, Tech Development, and Markets. This segmentation aligns with our qualitative analysis, which highlighted terms related to economic stability and financial activities. The SVM model's themes were Financial Metrics and Ratios, Corporate and Employment, and Market Analysis, reflecting a focus on business and economic performance.\\

For the non-peaceful word clouds, the LLM identified themes that starkly contrast with those of the peaceful countries. The Random Forest model's themes included Government Public, Legal Conflict, Geopolitical, Media, and Analysis Decision. Similarly, the SVM model's themes were Political and Social Issues, Legal Processes, Governance, Leadership, and Natural Resources. These themes reflect the conflict and instability prevalent in non-peaceful countries, with discussions focusing on political strife, legal battles, governance issues, and resource exploitation.\\ 

By comparing the themes identified by human analysts and the LLM, we can see a strong alignment, indicating that both human intuition and advanced AI systems recognize similar patterns in the linguistic features of peaceful and non-peaceful countries. This dual-method approach provides a comprehensive and validated understanding of the linguistic landscape in the context of peace and conflict studies.\\

\section*{Acknowledgment}

The data analyzed was generously provided by LexisNexis, downloaded and processed by Philippe Loustaunau, Vista Consulting LLC and who with Allegra Chen-Carrel, University of San Francisco identified the highest and lowest peace countries for supervised learning.  Exploratory analysis of the data was performed by Hongling Liu, Haoyue Qi, Xuanhao Wu, Yuxin Zhou, and Wenjie Zhu, as their Capstone Project for the the MS degree in Data Science at the Columbia University Data Science Institute. Further natural language processing and word counts were computed by Zach Stone and Natalia Wanda Zadrozna.  We thank Kate Sieck, Francine Chen, and Evelyn Kimani at the Toyota Research Foundation for their helpful suggestions and for funding the work to complete this project.



\end{document}